\documentclass{article}
\usepackage{spconf,amsmath,graphicx}
\usepackage{colortbl}
\usepackage{amsfonts}

\usepackage{multirow}

\usepackage{comment}
\usepackage{hyperref}
\usepackage{mathrsfs}
\usepackage{makecell}
\usepackage{txfonts}
\usepackage{dsfont}

\usepackage{etoolbox}
\apptocmd{\sloppy}{\hbadness 10000\relax}{}{}

\title{Domain-Generalized Face Anti-Spoofing with Unknown Attacks}

\name{Zong-Wei Hong$^{1}$, Yu-Chen Lin$^{1}$, Hsuan-Tung Liu$^{2}$, Yi-Ren Yeh$^{3}$, Chu-Song Chen$^{1}$
}
 \address{$^1$National Taiwan University \\
 $^{2}$E.SUN Financial Holding Co., Ltd.\\
 $^{3}$National Kaohsiung Normal University
 }
\usepackage[table,xcdraw]{xcolor}

\begin{document}
%
\maketitle
\begin{abstract}
Although face anti-spoofing (FAS) methods have achieved remarkable performance on specific domains or attack types, few studies have focused on the simultaneous presence of domain changes and unknown attacks, which is closer to real application scenarios. 
To handle domain-generalized unknown attacks, we introduce a new method, DGUA-FAS\footnote{ \href{https://github.com/AI-Application-and-Integration-Lab/DGUA_FAS}{\text{https://github.com/AI-Application-and-Integration-Lab/DGUA\_FAS}}}, which consists of a Transformer-based feature extractor and a synthetic unknown attack sample generator (SUASG).
The SUASG network simulates unknown attack samples to assist the training of the feature extractor.
Experimental results show that our method achieves superior performance on domain generalization FAS with known or unknown attacks.
\end{abstract}

\vspace{-11pt}
\section{Introduction}
\label{sec:intro}
With the widespread application of online payment, mobile device, and access control, FAS technology has received extensive attention.
However, when deploying FAS modules in real scenarios, we often need to tackle domain changes caused by different image sensors and photographing environments.
Besides, new attack types can occur and render well-trained systems ineffective.
It is thus crucial to conduct FAS methods that are robust to unseen domains and novel attacks.

Although traditional deep learning methods \cite{chen2019attention, yu2020searching} achieve great progress on specific datasets and protocols, they do not perform well on unseen domains 
and unknown attacks. 
Domain adaptation \cite{li2018unsupervised, liu2022source, zhou2022generative} and generalization techniques \cite{jia2020single, wang2022domain, wang2022patchnet, cai2022learning, liu2022causal, liao2023domain} have been developed to handle domain gaps. 
For handling novel attacks, zero/few-shot learning \cite{liu2019deep, qin2020learning} and anomaly detection \cite{arashloo2017anomaly, george2020learning} are used. 
However, tackling the simultaneous domain and attack changes is still challenging.

\par In this paper, we propose a method combing transformer-based architecture and open-set feature generator to simulate unknown attack samples scattered in the feature space. 
Our approach achieves favorable performance in several unseen domain and unknown attack settings. 
To our knowledge, this is the first time to introduce an open set feature generator as an attack sample generator for FAS.
Our design boosts the performance of the state-of-the-art transformer-based FAS model 
and we demonstrate that it is also effective based on other model backbones, 
indicating that our design can address 
the unseen domain and novel attacks for FAS effectively.

\begin{figure*}[t]
\centering\includegraphics[width=1.0\textwidth]{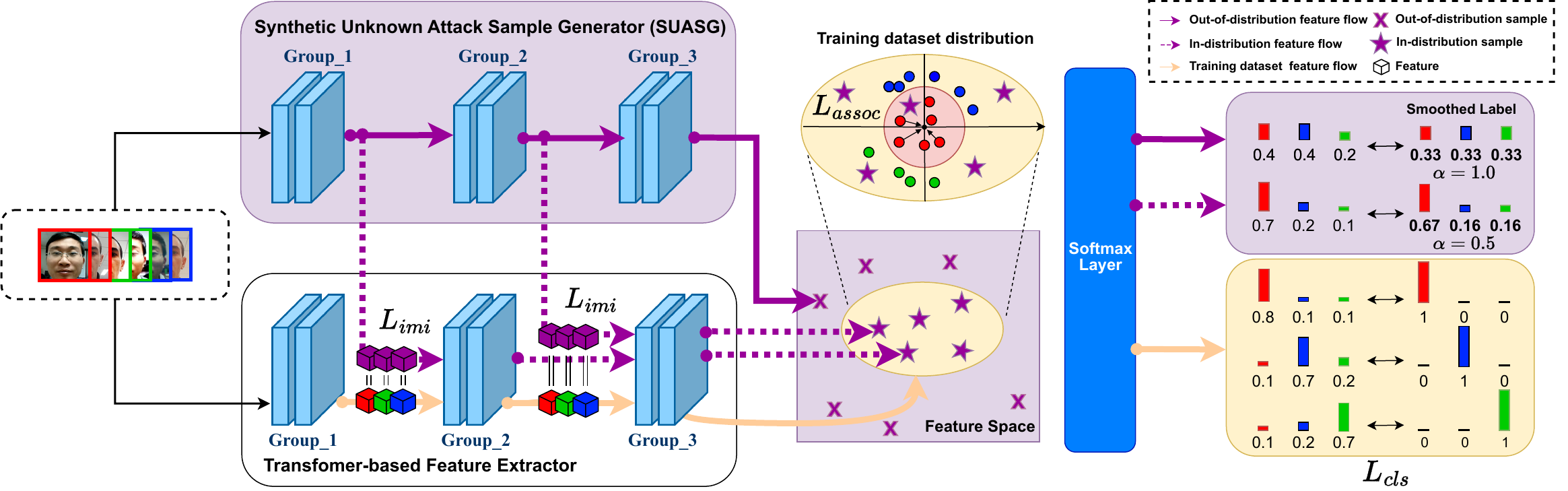}
\caption{Overall architecture of our proposed method (illustrated with 2 known attacks). Each training image with different types (\textit{Red} for real, \textit{Blue} for type-1, and \textit{Green} for type-2 attacks) is fed into both the SUASG and the Transformer-based Feature Extractor. There are two types of synthetic samples generated from each input, in-distribution samples (\textit{Purple Star}) and out-of-distribution samples (\textit{Purple Cross}). In-distribution samples are forced to be similar to those generated from corresponding groups of Feature Extractor by $\mathds{L}_{imi}$. Finally, we use $\textbf{Smoothed Label}$ as the objective of the output probabilities of synthetic samples; On the other side, real features generated by Feature Extractor are clustered at the origin of feature space by $\mathds{L}_{assoc}$, while features with the same attack type in different domains are forced to be in the same class by $\mathds{L}_{cls}$.
}
\label{fig:fig2}
\end{figure*}

\section{Related Work}
\label{sec:related}
 \noindent\textbf{Handling Cross Domain in FAS}.
 Compared with domain adaptation, domain generalization is more practical since no unseen domain data are needed in the training phase. 
 A typical technique 
 is to extract domain-independent features for classification. The approach in \cite{jia2020single} lets the real features from different domains be indistinguishable by single-side adversarial training, while spoofing features from each domain separate individually and all are far from real features through triplet learning. In \cite{wang2022domain}, two feature extractors are trained for content and style respectively, while content features are domain-invariant and the classification relies on style features. 
 The method of \cite{wang2022patchnet} formulates FAS as a patch-level classification problem which can utilize local features better. Patch features from the same image after non-distorted augmentation are forced to be invariant by the similarity loss, while the asymmetric angular margin softmax loss is used to enforce a larger margin with real-face features. 
 The approach of \cite{liao2023domain} extends the ideas of \cite{jia2020single}. 
 It uses concentration loss to centralize real face features at the origin and aggregate features of the same attack type from different domains.

\noindent\textbf{Handling Unknown Attacks in FAS}.
To tackle unknown attacks, previous studies introduce zero/few-shot learning and anomaly detection for FAS. 
In \cite{liu2019deep}, without semantic information of unknown attack types, the approach uses a deep tree structure to describe the known attacks and directly classifies the unknown attacks to the closest known attack type. 
The method in \cite{rostami2021detection} models the training distribution (including real and known attacks) by Gaussian Mixture Model (GMM), and draws samples out of the training distribution as unknown attack samples. 
As for anomaly detection, it assumes that real samples share more compact features 
than spoofs, and thus images can be classified by a one-class classifier instead of a traditional binary classifier, 
indicating that all the instances that do not belong to the only class are considered as attacks. 
Following this concept, 
a pair-wise one-class contrastive loss is used in \cite{george2020learning}
to build a 
one-class GMM. 
In addition, 
some 
studies formulate the unknown attack classification 
as an open set problem. Eg., in \cite{dong2021open}, Extreme Value Theorem (EVT) is adopted to detect unknown attacks. 
In our work, rather than using constraints for detection, we directly simulate spoofs as our open set data to 
provide more favorable results. 




\section{Methodology}
\label{sec:methodology}

We give the problem definition and then describe our method.
\subsection{Problem Formulation}
Suppose we have $M$ domains $\mathscr{D}=\{\mathcal{D}_{1:M}\}$ of the FAS training datasets; each dataset owns $|\mathcal{A}_i|$ types of attacks ($\mathcal{A}_{i=1:M}$) and one real face category $\mathcal{F}_i$.
In the training phase, the FAS predictor is learned by using only the training data of these known domains and types of attacks.

In the inference (testing) phase, the learned predictor is applied to some new domain (dataset) which contains all the original attacks $\bigcup\mathcal{A}_{i=1:M}$ and new types of attacks $\mathcal{A}_{unknown}$.
The purpose is to classify the input face as \textit{real} or \textit{spoof}, no matter the spoof data come from known or unknown attacks.

\subsection{Domain Generalized Unknown Attacks}
Our method (namely DGUA-FAS) handles both domain generalization and unknown attacks for FAS.

\noindent\textbf{Cross Domain.}
Classification loss ($\mathds{L}_{cls}$) 
and association loss ($\mathds{L}_{assoc}$) 
are the two main learning objectives 
for handling cross domain samples in DGUA-FAS.
We enforce the samples of the same  attack type to have similar (or different types to have dissimilar) embeddings regardless of their domains. 
Unlike the setting in~\cite{jia2020single} which encourages the same attacks in different domains to be separated either, 
we simply use the common multi-class cross-entropy loss for $\mathds{L}_{cls}$. 
Suppose we have $K$ types of attacks in the domain union $\bigcup\mathcal{A}_{i=1:M}$, the loss $\mathds{L}_{cls}$ 
separates the data into $K+1$ classes (Real face and $K$ attacks), no matter which domains the samples belong to.

In addition to separating the real and attack faces, we further enhance the data association among different domains.
Since the embeddings of real faces should be irrelevant to domains variations, 
we assume that all real samples from different datasets 
are similar and compact in the feature space.
Our real-face association loss follows 
\cite{liao2023domain}, which is designed as
\begin{equation}
\mathds{L}_{assoc} = \frac{1}{\left|F_{real}\right|} \sum_{f \in F_{real}}\left\|f\right\|_1, \forall f \in F_{real},
\end{equation}
where $F_{real}$ is the set of 
real-face feature embeddings, and 
they are enforced to concentrate at the origin of feature space. 

DGUA-FAS is trained with the loss $\mathds{L}_{main}=\mathds{L}_{cls}+\lambda\cdot\mathds{L}_{assoc}$ together with some other complementary loss terms that are introduced below.
Without loss of generality, we use a transformer as the backbone network (yet CNNs can be used as well),
which is shown as the Transformer-based Feature Extractor in Fig.~\ref{fig:fig2} (lower-left part).


\noindent\textbf{Unknown Attack}. 
We have the data only in the known attacks $\mathcal{A}_{i=1:M}$ for training but do not have the unknown attack data in $\mathcal{A}_{unknown}$.
To make our model 
capable of handling open set samples, 
an intuitive idea is to add some simulated unknown-attack samples in the training process. 
In \cite{rostami2021detection}, only the samples out of the training dataset distribution are produced as novel attacks. 
However, some unknown attacks could also be overlapped with the training distribution. 
Ignoring the possibility of the within-distribution similarity often leads to unreliable predictions.
Since our approach learns a classifier of $K+1$ classes for prediction, we hope that the simulated samples of unknown attacks can be evenly distributed over the $K+1$ regions.
E.g., considering $K=2$ attack types in Fig.~\ref{fig:fig2}, we would hope the output probability of the unknown attacks to be approximately 0.33 for the 3 classes (upper-right part).
Hence, we simulate not only the out-of-distribution but also the in-distribution data. 

To achieve this, we leverage the different layers in a network to generate samples of different difficulty levels, so that they can scatter both inside and outside of the known-classes distribution.
Inspired by DiAS \cite{moon2022difficulty}, we complement an architecture called Synthetic Unknown Attack Sample Generator (SUASG) to produce synthetic unknown samples for FAS during training (upper-left of Fig.~\ref{fig:fig2}).
SUASG shares the same structure as the Transformer-based Feature Extractor but has its own weights. 
To train SUASG, We first divide both networks into three groups. 
Each training input image is fed into both networks.
Between the corresponding groups, we add the following loss (called \textit{imitation loss}) to constrain the features generated by the two networks:
\begin{equation}
\mathds{L}_{imi}=\frac{1}{N} \sum_{i=1}^N \sum_{g=1}^2 \|f_{SUASG}^{g,i} - f_{extract}^{g,i}\|_1, 
\label{eq:imitation}
\end{equation}
with $N$ the mini-batch size;$f_{SUASG}^{g,i}$ and $f_{extract}^{g,i}$ are the features of the $i$-th image extracted by the $g$-th groups of SUASG and Transformer-based Feature Extractor, respectively. 
To train the network, an input image $i$ passes through a total of $G$ paths ($G=3$ groups in our case).
One is the path passing through only SUASG.
Each of the remaining paths goes through the 1st to the $g$-th groups of SUASG and then the $(g+1)$-th to the $G$-th groups of the Transformer-based Feature Extractor $(1\leq g<G)$.
As for the path through SUASG only, the generated samples are not constrained by their final group, which we consider to be a simulation of out-of-distribution samples.
When $1\leq g<G$, the samples produced are constrained by the imitation losses 
and then share the same final groups of Transformer-based Feature Extractor. 
We then use them to simulate the in-distribution data.
Hence, both the out-of-distribution and in-distribution samples of unknown classes are generated.
We empirically find that it achieves favorable results in dealing with unknown attacks more efficiently.



\begin{table}[htb]
\caption{Datasets used in our evaluation. (i) and (v) denote ‘images’ and ‘videos’, respectively.}

\resizebox{1.0\linewidth}{!}{
\begin{tabular}{|c|c|c|}
\hline
 
Dataset       & Number of videos/images & Attack type                                                                                                               \\ \hline
 
OULU-NPU\cite{boulkenafet2017oulu}      & 3600 (v)                 & print, replay                                                                                                             \\ \hline
 
CASIA-FASD\cite{6199754}    & 600 (v)                  & print, replay                                                                                                             \\ \hline
 
MSU-MFSD\cite{wen2015face}      & 280 (v)                  & print, replay                                                                                                             \\ \hline
 
Replay-Attack \cite{chingovska2012effectiveness} & 1200 (v)                 & print, replay                                                                                                             \\ \hline
CelebA-Spoof \cite{CelebA-Spoof} & 625537 (i)               & print, replay, paper mask                                                                         \\ \hline
 
WMCA     \cite{george2019biometric}     & 1679 (v)                 & \begin{tabular}[c]{@{}c@{}}print, replay, Partial (glasses), \\ Mask (plastic, silicone, and paper, Mannequin)\end{tabular} \\ \hline
\end{tabular}}
\label{table:datasets' info}
\vspace{-10pt}
\end{table}

We implement the training process by iterating the following two steps.
In the first step, we fix the Transformer-based Feature Extractor and train SUASG by the imitation loss (Eq.~\ref{eq:imitation}) and $\mathds{L}_{cls}$. 
In the second step, we fix SUASG and train the Transformer-based Feature Extractor by minimizing 
\begin{equation}
\mathds{L}_{extract} = \mathds{L}_{main} + \mathds{L}_{cls}(x_{SID},\hat{y}_{SID}) + \mathds{L}_{cls}(x_{SOOD}, \hat{y}_{SOOD}),
\end{equation}
where $x_{SID}$ and $x_{SOOD}$ are the output probabilities of the synthetic in-distribution and out-of-distribution samples, respectively, and $\hat{y}_{SID}$ and $\hat{y}_{SOOD}$ are their respective target probabilities.  $\mathds{L}_{cls}$ is the multi-class cross-entropy loss.
As mentioned, we hope the unknown-class samples to be equally distributed as [$1/(K+1)$] over the $K+1$ classes.
On the other hand, as the simulated data are synthesized from the original training data, the original labels could serve as the pseudo labels for the synthesized data.
We thus 
smooth them to form the target probabilities of the synthetic data as suggested by \cite{moon2022difficulty}:
\begin{equation}
\hat{y} = (1 - \alpha) \cdot y + \alpha/(K+1) \cdot \textbf{u},
\end{equation}
with $y$ the original label of the input image $x$, and $\textbf{u}$ is the all-one vector.
A higher $\alpha$ is set for the out-of-distribution synthetic data and vice versa.
When training is finished, in the inference stage, only the Transformer-based Feature Extractor together with the final classification layer are used for testing.

\section{Experiments}
\label{sec:Experiment}

To verify our DGUA-FAS, we first examine its performance on the domain-generalized problem. Then, we present the results on the problem of unseen domains with both known and unknown attacks.


\noindent\textbf{Datasets and Metrics}. The datasets CASIA-FASD \cite{6199754} (C), MSU-MFSD \cite{wen2015face} (M), Idiap Replay-attack \cite{chingovska2012effectiveness} (I),  OULU-NPU \cite{boulkenafet2017oulu} (O), CelebA-Spoof \cite{CelebA-Spoof}, and WMCA\cite{george2019biometric} are used to evaluate our method.
The first four 
(C\&M\&I\&O) include only print and replay attacks, while the rest (i.e., CelebA-Spoof \& WMCA) contain more diverse attack types such as glasses, silicone masks, and paper masks 
(Table \ref{table:datasets' info}). 
Following previous works, we utilize the Half Total Error Rate (HTER) and the Area Under Curve (AUC) as the evaluation metrics.

\noindent\textbf{Implemetation detail}.
We use MobileViT-S as our backbone network. 
The version 
we use is \cite{mehta2022mobilevit} and the model is pre-trained on ImageNet-1K. 
We choose Adam optimizer and let the learning rate and weight decay parameter to be $10^{-4}$ and $10^{-6}$. 
We conduct all experiments on a single RTX 3090 GPU and set $\lambda=1.0$, 
$\alpha= 0.5$ and $\alpha=1.0$ for in-distribution and out-of-distribution samples respectively. 
To divide the SUASG network and yield the synthetic samples, we consider $conv_1$ and $layer_1$ as the first group, $layer_2$ and $layer_3$ as the second group, and $layer_4$, $layer_5$ as the third group.

\begin{table}[t]
\caption{AUC (\%) of the proposal method and previous domain-generalized methods on leave-one-out Setting. The best results are \textbf{bolded}, and the second best 
is \underline{underlined}.}

\resizebox{1.0\linewidth}{!}
{
\begin{tabular}{|c|c|c|c|c|c|}
\hline
 \textbf{Method} & \multicolumn{1}{c|}{\makecell{ O \& C \& I \\ to M}}              & \multicolumn{1}{c|}{\makecell{O \& M \& I \\ to C}}              & \multicolumn{1}{c|}{ \makecell{ O \& C \& M \\ to I }}               & \multicolumn{1}{c|}{\makecell{I \& C \& M \\to O}}                & \multicolumn{1}{c|}{\makecell{Average \\AUC (\%)}}                           \\ \hline

SSDG-R \cite{jia2020single}         & 97.17                & 95.94                 & 96.59                 & 91.54              & 95.31           \\

SSAN-R \cite{wang2022domain}        &  \underline{98.75}   & { 96.67}              & 96.79                 & 93.63              & 96.46           \\

PatchNet \cite{wang2022patchnet}  & 98.46                & 94.58                 & 95.67                 & \underline{95.07}           & 95.945          \\
HFN+MP \cite{cai2022learning}  & 97.28                & 96.09                 & 90.67                 &  94.26           & 94.575          \\
CIFAS \cite{liu2022causal}  & 96.32                & 95.30                 & 97.24                 &  93.44           & 95.575          \\
DiVT-M \cite{liao2023domain}      & \textbf{99.14}       & \underline{96.92}     & \textbf{99.29}        & 94.04              & \underline{97.347}          \\
\hline
Proposed approach                 & 98.156               & \textbf{97.0}         &  \underline{99.187}               & \textbf{96.369}       & \textbf{97.678} \\ \hline
\end{tabular}}

\label{tab:crossv2}
\vspace{-15pt}
\end{table}

\noindent\textbf{Results on Domain-generalized Settings}.
First, we show the results of DGUA-FAS on the traditional domain-generalized settings
, where the unseen domain has no new types of attacks.
The purpose is to verify that our method, though can handle new attacks on new domains, still perform well on the standard domain-generalized scenario without sacrificing the performance.
As shown in Table \ref{tab:crossv2} (leave-one-out setting) and Table \ref{Table:limited} (limited-source setting) on the C\&M\&I\&O datasets, our method is competitive with the SOTA methods 
even though we have used further synthetic unknown attack samples during training. 
Table \ref{Table:limited} shows that our performance even exceeds those of the SOTA methods. 
We speculate that, in this case, synthetic samples enrich limited training data by augmenting the training distribution to better fit the testing distribution, thus leading to more favorable results. 

In addition, we also change the backbone of our approach to CNNs and examine its performance.
We use ResNet18 (also used in SSDG \cite{jia2020single}) to build our model, and the main losses follow those in the settings of \cite{jia2020single}.
We use layers 1-5 as the first, 6-13 as the second, and the rest as the final groups in ResNet18. As shown in Table \ref{Table:limited}, our method significantly improves SSDG in the limited-source setting, revealing that our framework can 
work well with both transformer 
and 
CNN-based models. 






\vspace{-6pt}
\begin{table}[htb]
\caption{The experimental results on the comparison between our method and SOTA methods on the limited-source setting. }

\label{Table:limited}
\resizebox{1.0\linewidth}{!}{
\begin{tabular}{|c |c c| cc|}
\hline
\multirow{2}{*}{\textbf{Method}}              & \multicolumn{2}{c|}{M\&I to C}                         & \multicolumn{2}{c|}{M\&I to O}                         \\ \cline{2-5} 
& \multicolumn{1}{c|}{HTER (\%)}       & \multicolumn{1}{c|}{AUC (\%)}        & \multicolumn{1}{c|}{HTER (\%)}       &  \multicolumn{1}{c|}{AUC (\%)}        \\ \hline

SSAN-R \cite{wang2022domain}                                  & \multicolumn{1}{c|}{25.56}           & 83.89           & \multicolumn{1}{c|}{24.44}           & 82.86           \\ 

HFN+MP \cite{cai2022learning}                                  & \multicolumn{1}{c|}{30.89}           & 72.48           & \multicolumn{1}{c|}{20.94}           & 86.71           \\ 

CIFAS \cite{liu2022causal}                                   & \multicolumn{1}{c|}{22.67}           & 83.89           & \multicolumn{1}{c|}{24.63}           & 81.48           \\ 

DiVT-M \cite{liao2023domain}                                 & \multicolumn{1}{c|}{20.11}           & 86.71          & \multicolumn{1}{c|}{23.61}          & 85.73          \\ 
SSDG-R \cite{jia2020single}                                  & \multicolumn{1}{c|}{19.86}           & 86.46           & \multicolumn{1}{c|}{27.92}           & 78.72           \\ \hline

Proposed approach w/ setting of SSDG \cite{jia2020single}                                         & \multicolumn{1}{c|}{\textbf{18.667}} & \textbf{87.089} & \multicolumn{1}{c|}{{\underline{ 20.139}}}    & {\underline {87.523}}    \\ 

Proposed approach                                         & \multicolumn{1}{c|}{\underline{19.222}}    & \underline{86.806}    & \multicolumn{1}{c|}{\textbf{20.052}} & \textbf{88.746} \\ \hline
\end{tabular}}
\vspace{-6pt}
\end{table}

\noindent\textbf{Results on Unseen Domain with Unknown Attacks}.
Previous works mainly tackle the scenario of new attacks on the same domain.
There are still very few studies (\cite{liu2022source,liu2022causal}) which have reported the results on the cross-domain FAS with both known and unknown types of attacks in the testing phase.
Following the setting in \cite{liu2022source,liu2022causal}, we compare our method with them on training with M\&C\&O and testing on CelebA-Spoof datasets.
We also do the experiments on training with O\&M\&I\&C and testing on WMCA datasets, and reproduce the recent domain-generalized method \cite{liao2023domain} on both settings for comparison.
Both setups of M\&C\&O to CelebA-Spoof and O\&M\&I\&C to WMCA contain only print and replay attacks during training, while unknown attacks like glasses and masks appear in the testing phase additionally. 

The results are shown in Table \ref{Table:unseen attacks}. 
As can be seen, our method outperforms CIFAS~\cite{liu2022causal} on the M\&C\&O to CelebA-Spoof setting, but performs worse than \cite{liu2022source}.
It is because \cite{liu2022source} is a domain-adaptation rather than a domain-generalized approach; a lot of unlabeled data in the new domain are collected and allowed for training in \cite{liu2022source}.
However, in real scenarios, we could not gather so much new-attack data for building our model in practice.
As for the comparison on domain generalization with unknown attacks, our method performs more favorably than \cite{liu2022causal} and \cite{liao2023domain}, revealing that the proposed solution is effective to tackle the domain generalized FAS containing both known and unknown attacks.


\begin{table}[t]
\caption{Results on 
new domain with novel attacks. (*) the method uses further the unlabeled data in the new domain.
} 
\vspace{1.5pt}
\label{Table:unseen attacks}
\centering
\resizebox{1.0\linewidth}{!}{
\begin{tabular}{| c | c c| cc|}
\hline
\multirow{2}{*}{\textbf{Method}} & \multicolumn{2}{c|}{\begin{tabular}[c]{@{}c@{}}M\&C\&O to \\ CelebA-Spoof\end{tabular}} & \multicolumn{2}{c|}{\begin{tabular}[c]{@{}c@{}}I\&M\&C\&O to \\ WMCA\end{tabular}} \\ \cline{2-5} 
 & \multicolumn{1}{c|}{HTER (\%)}                        & AUC (\%)                        & \multicolumn{1}{c|}{HTER (\%)}         & AUC (\%)          \\ \hline
*SDA-FAS \cite{liu2022source}                                & \multicolumn{1}{c|}{\textbf{18.9}}                    & \textbf{90.9}                   & \multicolumn{1}{c|}{\textbf{-}}        & -                                         \\ 
CIFAS \cite{liu2022causal}                                   & \multicolumn{1}{c|}{24.6}                             & 83.2                            & \multicolumn{1}{c|}{-}                 & -                                         \\ 
DiVT-M \cite{liao2023domain}                               & \multicolumn{1}{c|}{25.075}                           & 82.335                          & \multicolumn{1}{c|}{\underline{22.364}}            & \underline{86.816}            \\ \hline
Proposed approach                                         & \multicolumn{1}{c|}{\underline{ 21.442}}                     & \underline{ 86.351}                    & \multicolumn{1}{c|}{\textbf{20.624}}   & \textbf{88.071}   \\ \hline
\end{tabular}}
\vspace{-12pt}
\end{table}

\begin{table}[t]
\caption{Ablation study on using different levels of synthetic samples on the leave-one-out domain generalized setting. 
}
\vspace{1.5pt}
\label{Table:ablation}
\centering
\resizebox{0.9\linewidth}{!}{
\begin{tabular}{|c|c|cc|}
\hline
\multirow{2}{*}{\makecell{In-distribution \\samples}}  &   \multirow{2}{*}{\makecell{Out-of-distribution \\samples}}                        & \multicolumn{2}{c|}{Average}      \\ \cline{3-4} 
 &  & \multicolumn{1}{c|}{HTER (\%)}         & AUC (\%)           \\ \hline
-                                                      & -                                                      & \multicolumn{1}{c|}{9.28325}          & 96.48425          \\ 
v                                                      & -                                                      & \multicolumn{1}{c|}{7.97525}           & 97.3725          \\ 
-                                                      & v                                                      & \multicolumn{1}{c|}{8.4375}           & 97.07925          \\ 
v                                                      & v                                                      & \multicolumn{1}{c|}{\textbf{7.1}} & \textbf{97.678} \\ 
\hline
\end{tabular}}
\vspace{-10pt}
\end{table}

\noindent\textbf{Ablation Study}.
We conduct the experiments on the effectiveness of in-distribution samples and out-of-distribution samples.
The results of using (or not using) in-distribution and out-of-distribution synthetic samples are reported in Table \ref{Table:ablation}.
The results reveal that the simulated samples can effectively 
improve the performance alone, and the best performance is achieved when using both.


\vspace{-10pt}
\section{Conclusion}
\label{sec:conclusion}
We introduce DGUA-FAS, a simple yet effective approach to handle FAS problems with both known and unknown attacks in a new domain.
Our method centralizes the real face and separates the attack types in the embedding space regardless of the domains.
It leverages a simulated-data generator to produce in-distribution and out-of-distribution samples of the unknown classes for training.
Experimental results show that DGUA-FAS is not only effective for traditional domain-generalized FAS but can also achieve the most favorable performance on the FAS tasks with additional unknown attacks in the new domain.
In the future, we plan to extend our method to video-based approaches.


\noindent\textbf{Acknowledgement}. This work was supported in part by E.SUN Financial Holding, and National Science and Technology Council in Taiwan (NSTC 111-2634-F-006-012, NSTC 111-2634-F-002-023). Computational and storage resources are partly supported from NCHC of NARLabs in Taiwan.


\renewcommand{\refname}{}

\vfill\pagebreak

\clearpage

\let\oldthebibliography\thebibliography
\let\endoldthebibliography\endthebibliography
\renewenvironment{thebibliography}[1]{
  \begin{oldthebibliography}{#1}
    \setlength{\itemsep}{0.2em}
    \setlength{\parskip}{0em plus 2pt}
}
{
  \end{oldthebibliography}
}

\bibliographystyle{IEEEbib}

\bibliography{strings,refs}

\begin{thebibliography}{10}

\bibitem{chen2019attention}
Haonan Chen, Guosheng Hu, Zhen Lei, Yaowu Chen, Neil~M Robertson, and Stan~Z
  Li,
\newblock ``Attention-based two-stream convolutional networks for face spoofing
  detection,''
\newblock {\em IEEE TIFS}, vol. 15, pp. 578--593, 2019.

\bibitem{yu2020searching}
Zitong Yu, Chenxu Zhao, Zezheng Wang, Yunxiao Qin, Zhuo Su, Xiaobai Li, Feng
  Zhou, and Guoying Zhao,
\newblock ``Searching central difference convolutional networks for face
  anti-spoofing,''
\newblock in {\em CVPR}, 2020, pp. 5295--5305.

\bibitem{li2018unsupervised}
Haoliang Li, Wen Li, Hong Cao, Shiqi Wang, Feiyue Huang, and Alex~C Kot,
\newblock ``Unsupervised domain adaptation for face anti-spoofing,''
\newblock {\em IEEE TIFS}, vol. 13, no. 7, pp. 1794--1809, 2018.

\bibitem{liu2022source}
Yuchen Liu, Yabo Chen, Wenrui Dai, Mengran Gou, Chun-Ting Huang, and Hongkai
  Xiong,
\newblock ``Source-free domain adaptation with contrastive domain alignment and
  self-supervised exploration for face anti-spoofing,''
\newblock in {\em ECCV}. Springer, 2022, pp. 511--528.

\bibitem{zhou2022generative}
Qianyu Zhou, Ke-Yue Zhang, Taiping Yao, Ran Yi, Kekai Sheng, Shouhong Ding, and
  Lizhuang Ma,
\newblock ``Generative domain adaptation for face anti-spoofing,''
\newblock in {\em ECCV}. Springer, 2022, pp. 335--356.

\bibitem{jia2020single}
Yunpei Jia, Jie Zhang, Shiguang Shan, and Xilin Chen,
\newblock ``Single-side domain generalization for face anti-spoofing,''
\newblock in {\em CVPR}, 2020, pp. 8484--8493.

\bibitem{wang2022domain}
Zhuo Wang, Zezheng Wang, Zitong Yu, Weihong Deng, Jiahong Li, Tingting Gao, and
  Zhongyuan Wang,
\newblock ``Domain generalization via shuffled style assembly for face
  anti-spoofing,''
\newblock in {\em CVPR}, 2022, pp. 4123--4133.

\bibitem{wang2022patchnet}
Chien-Yi Wang, Yu-Ding Lu, Shang-Ta Yang, and Shang-Hong Lai,
\newblock ``Patchnet: A simple face anti-spoofing framework via fine-grained
  patch recognition,''
\newblock in {\em CVPR}, 2022, pp. 20281--20290.

\bibitem{cai2022learning}
Rizhao Cai, Zhi Li, Renjie Wan, Haoliang Li, Yongjian Hu, and Alex~C Kot,
\newblock ``Learning meta pattern for face anti-spoofing,''
\newblock {\em IEEE TIFS}, vol. 17, pp. 1201--1213, 2022.

\bibitem{liu2022causal}
Yuchen Liu, Yabo Chen, Wenrui Dai, Chenglin Li, Junni Zou, and Hongkai Xiong,
\newblock ``Causal intervention for generalizable face anti-spoofing,''
\newblock in {\em ICME}. IEEE, 2022, pp. 01--06.

\bibitem{liao2023domain}
Chen-Hao Liao, Wen-Cheng Chen, Hsuan-Tung Liu, Yi-Ren Yeh, Min-Chun Hu, and
  Chu-Song Chen,
\newblock ``Domain invariant vision transformer learning for face
  anti-spoofing,''
\newblock in {\em WACV}, 2023, pp. 6098--6107.

\bibitem{liu2019deep}
Yaojie Liu, Joel Stehouwer, Amin Jourabloo, and Xiaoming Liu,
\newblock ``Deep tree learning for zero-shot face anti-spoofing,''
\newblock in {\em CVPR}, 2019, pp. 4680--4689.

\bibitem{qin2020learning}
Yunxiao Qin, Chenxu Zhao, Xiangyu Zhu, Zezheng Wang, Zitong Yu, Tianyu Fu, Feng
  Zhou, Jingping Shi, and Zhen Lei,
\newblock ``Learning meta model for zero-and few-shot face anti-spoofing,''
\newblock in {\em AAAI}, 2020, vol.~34, pp. 11916--11923.

\bibitem{arashloo2017anomaly}
Shervin~Rahimzadeh Arashloo, Josef Kittler, and William Christmas,
\newblock ``An anomaly detection approach to face spoofing detection: A new
  formulation and evaluation protocol,''
\newblock {\em IEEE access}, vol. 5, pp. 13868--13882, 2017.

\bibitem{george2020learning}
Anjith George and S{\'e}bastien Marcel,
\newblock ``Learning one class representations for face presentation attack
  detection using multi-channel convolutional neural networks,''
\newblock {\em IEEE TIFS}, vol. 16, pp. 361--375, 2020.

\bibitem{rostami2021detection}
Mohammad Rostami, Leonidas Spinoulas, Mohamed Hussein, Joe Mathai, and Wael
  Abd-Almageed,
\newblock ``Detection and continual learning of novel face presentation
  attacks,''
\newblock in {\em ICCV}, 2021, pp. 14851--14860.

\bibitem{dong2021open}
Xin Dong, Hao Liu, Weiwei Cai, Pengyuan Lv, and Zekuan Yu,
\newblock ``Open set face anti-spoofing in unseen attacks,''
\newblock in {\em ACM MM}, 2021, pp. 4082--4090.

\bibitem{moon2022difficulty}
WonJun Moon, Junho Park, Hyun~Seok Seong, Cheol-Ho Cho, and Jae-Pil Heo,
\newblock ``Difficulty-aware simulator for open set recognition,''
\newblock in {\em ECCV}. Springer, 2022, pp. 365--381.

\bibitem{boulkenafet2017oulu}
Zinelabinde Boulkenafet, Jukka Komulainen, Lei Li, Xiaoyi Feng, and Abdenour
  Hadid,
\newblock ``Oulu-npu: A mobile face presentation attack database with
  real-world variations,''
\newblock in {\em FG 2017}. IEEE, 2017, pp. 612--618.

\bibitem{6199754}
Zhiwei Zhang, Junjie Yan, Sifei Liu, Zhen Lei, Dong Yi, and Stan~Z. Li,
\newblock ``A face antispoofing database with diverse attacks,''
\newblock in {\em ICB}, March 2012, pp. 26--31.

\bibitem{wen2015face}
Di~Wen, Hu~Han, and Anil~K Jain,
\newblock ``Face spoof detection with image distortion analysis,''
\newblock {\em IEEE TIFS}, vol. 10, no. 4, pp. 746--761, 2015.

\bibitem{chingovska2012effectiveness}
Ivana Chingovska, Andr{\'e} Anjos, and S{\'e}bastien Marcel,
\newblock ``On the effectiveness of local binary patterns in face
  anti-spoofing,''
\newblock in {\em BIOSIG}. IEEE, 2012, pp. 1--7.

\bibitem{CelebA-Spoof}
Yuanhan Zhang, Zhenfei Yin, Yidong Li, Guojun Yin, Junjie Yan, Jing Shao, and
  Ziwei Liu,
\newblock ``Celeba-spoof: Large-scale face anti-spoofing dataset with rich
  annotations,''
\newblock in {\em ECCV}, 2020.

\bibitem{george2019biometric}
Anjith George, Zohreh Mostaani, David Geissenbuhler, Olegs Nikisins, Andr{\'e}
  Anjos, and S{\'e}bastien Marcel,
\newblock ``Biometric face presentation attack detection with multi-channel
  convolutional neural network,''
\newblock {\em IEEE TIFS}, vol. 15, pp. 42--55, 2019.

\bibitem{mehta2022mobilevit}
Sachin Mehta and Mohammad Rastegari,
\newblock ``Mobilevit: Light-weight, general-purpose, and mobile-friendly
  vision transformer,''
\newblock in {\em ICLR}, 2022.

\end{thebibliography}

\end{document}